\documentclass[11pt]{article}

\usepackage[final]{latex/acl}
\usepackage{times}
\usepackage{latexsym}

\usepackage[T1]{fontenc}

\usepackage[utf8]{inputenc}

\usepackage{microtype}

\usepackage{inconsolata}

\usepackage{graphicx}

\usepackage[T1]{fontenc}

\usepackage{amsfonts}
\usepackage{amssymb}

\usepackage{times}
\usepackage{latexsym}
\usepackage{booktabs}
\usepackage{multirow}
\usepackage{mathtools}
\usepackage{etoolbox,xspace}
\usepackage[title,page]{appendix}

\usepackage[inline,shortlabels]{enumitem}
\usepackage{amsmath}
\usepackage{csquotes}

\usepackage[utf8]{inputenc}

\usepackage{microtype}

\usepackage{subcaption}
\usepackage{adjustbox}

\usepackage{colortbl}

\usepackage{graphicx}

\usepackage[capitalise]{cleveref}

\usepackage{tikz}

\crefname{figure}{Fig.}{Figs.}
\crefname{equation}{}{}
\crefdefaultlabelformat{#2\textup{#1}#3}

\newcolumntype{C}[1]{>{\columncolor{white}}p{#1}}

\input{latex/macros}

\title{Talking to Yourself: Defying Forgetting in Large Language Models}

\author{
 \textbf{Yutao Sun\textsuperscript{1}},
 \textbf{Mingshuai Chen\textsuperscript{1}},
 \textbf{Tiancheng Zhao\textsuperscript{2,3}},\\
 \textbf{Phillip Miao\textsuperscript{4}},
 \textbf{Zilun Zhang\textsuperscript{1}},
 \textbf{Haozhan Shen\textsuperscript{1}},
 \textbf{Ruizhe Zhu\textsuperscript{5}},
 \textbf{Jianwei Yin\textsuperscript{1}}
\\
\\
 \textsuperscript{1}Zhejiang University,
 \textsuperscript{2}Binjiang Institute of Zhejiang University,
 \textsuperscript{3}Om AI Research, \\
 \textsuperscript{4}Stanford University,
 \textsuperscript{5}ETH Zürich,
\\
 \small{
   \textbf{Correspondence:} \href{mailto:m.chen@zju.edu.cn}{m.chen@zju.edu.cn}, \href{mailto:tianchez@zju-bj.com}{tianchez@zju-bj.com}, \href{mailto:zjuyjw@zju.edu.cn}{zjuyjw@zju.edu.cn}
 }
}

\begin{document}
\maketitle

\begin{abstract}

Catastrophic forgetting remains a major challenge when fine-tuning large language models (LLMs) on narrow, task-specific data, often degrading their general knowledge and reasoning abilities. We propose \methodname, a lightweight self-augmentation routine in which an LLM generates self-dialogues prior to fine-tuning, and the resulting self-authored data are mixed with task data without modifying optimization or training schedules.

Despite requiring no external data or additional tuning, \methodname consistently mitigates catastrophic forgetting while improving in-domain performance. Across 50 evaluation scenarios, it maintains performance comparable to the original model and achieves the best results in 40 cases, outperforming common baselines such as layer freezing and external data mixing. Guided by these empirical findings, we further present a theoretical analysis suggesting that forgetting can partly stem from style-induced parameter drift, and that self-alignment through self-generated data provides an effective means to counteract this effect. Overall, our results indicate that self-augmentation offers a simple and effective mechanism for robust LLM adaptation without incurring catastrophic forgetting.

\end{abstract}

\section{Introduction}

Large language models (LLMs) now underpin modern human machine intelligence, driving agents that code, reason, converse, and assist across domains \cite{nam2024using,roziere2023code,DBLP:conf/aaai/AnandPKNNJS25}. Their hallmark strength—an ability to rapidly \emph{specialize} when fine-tuned on domain-specific data—has also exposed a critical weakness: each specialization step risks overwriting the general reasoning, linguistic, and world knowledge that make these models broadly useful in the first place \cite{DBLP:journals/corr/abs-2308-08747,kemker2018measuring,zhang2024preserving,ramasesh2021effect}. This phenomenon, known as \emph{catastrophic forgetting}, undermines the long-term stability and trustworthiness of LLM-driven agents—e.g., a chatbot fine-tuned for customer support may suddenly lose arithmetic competence, factual recall, or safety alignment, compromising reliability in real-world interaction.

\begin{figure}[t]
  \centering
  \includegraphics[width=\linewidth]{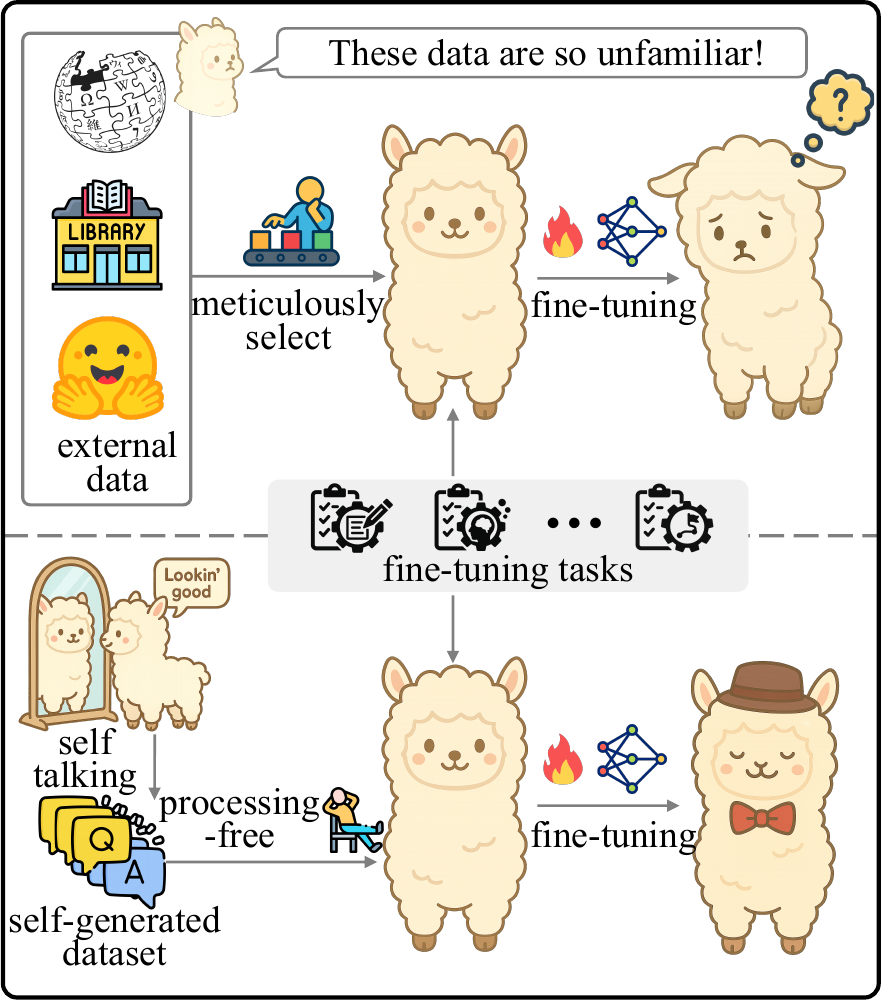}
  \caption{External rehearsal (top) fine-tunes on unfamiliar data and risks forgetting, whereas \methodname (bottom) lets the model “talk to itself,” mixing a processing-free, self-generated corpus with task data to retain prior skills.}
  \label{fig:intro}
\end{figure}

Existing safeguards are either expensive or fragile.
One family of methods relies on \emph{data rehearsal}: practitioners curate or purchase large, heterogeneous corpora to “remind’’ the model of what it once knew \cite{DBLP:conf/acl/HuangCWYLSYS24,mok2023large,de2019episodic}.
Another family introduces explicit regularizers—such as Elastic Weight Consolidation \cite{DBLP:conf/nips/XiangTGSWYH23} or Kullback–Leibler penalties \cite{nikoloutsopoulos2024kullback}—that require per-task tuning and often yield unstable results.
In practice, however, such parameter-based regularization remains uncommon in large-scale SFT pipelines: estimating and storing parameter importance (e.g., via Fisher information) for billions of weights incurs prohibitive memory and communication cost, making these methods impractical beyond small or controlled setups.
As a result, most real-world adaptations resort to cheaper heuristics—data mixing \cite{DBLP:journals/csur/CaoZDWZ25}, layer freezing \cite{DBLP:conf/iclr/ZhengCQ025}, or LoRA \cite{DBLP:conf/iclr/HuSWALWWC22}—yet these too struggle to maintain general ability over long training horizons.



In this paper, we ask a different question: \emph{What if the model could rehearse on data of its own making?}
Inspired by self-instruction frameworks such as \textsc{Self-Instruct} \cite{selfinstruct}, \textsc{Magpie} \cite{DBLP:conf/iclr/XuJNDP0L25}, and \textsc{Crescent} \cite{DBLP:journals/corr/abs-2502-13441}, we propose \textbf{Self-Augmented Supervised Fine-Tuning} (\methodname)—a lightweight self-modeling routine that lets an LLM generate and integrate its own rehearsal data before adaptation.
Before touching any downstream corpus, the frozen base model conducts short self-talking, as shown in \cref{fig:intro}.
The resulting self-authored dataset is folded into the normal fine-tuning stream without changing optimization or schedules.


Despite its stark simplicity—no external data, no extra losses, and no hyper-tuning—\methodname consistently stabilizes learning and enhances specialization.
Across five heterogeneous tasks drawn from \textsc{Super-Natural-Instructions} \cite{DBLP:conf/emnlp/WangMAKMNADASPK22}, \methodname restores general-ability performance in \textbf{all 50} evaluation scenarios and achieves the best overall results in \textbf{40} of them, outperforming mainstream strategies such as layer freezing \cite{DBLP:conf/iclr/ZhengCQ025} and external data mixing while achieving the most balanced and stable trade-off between retention and adaptation.
Unlike prevailing fine-tuning practices that require extensive effort to curate or match suitable external datasets—an increasingly difficult task given that most large models do not disclose their pretraining corpora.
We further provide a theoretical explanation showing that this \methodname process mitigates style-induced parameter drift—the underlying cause of spurious forgetting—thereby enabling a form of autonomous, self-consistent data modeling.


Our contributions are threefold:

\begin{enumerate}
\item \textbf{A self-modeling data paradigm.} We present \methodname, a forgetting-mitigation framework that transforms rehearsal into a form of autonomous data modeling—requiring no external corpus, task-specific filtering, or additional loss design.
\item \textbf{Comprehensive validation.} Extensive experiments and analyses across five benchmarks and multiple architectures confirm that self-generated data can match or surpass curated external datasets in retaining general skills.
\item \textbf{Mechanistic and theoretical insight.} We identify and formalize \emph{style-induced forgetting} as a key failure mode and show theoretically that self-aligned data suppresses these harmful stylistic gradients, yielding stable, interpretable adaptation dynamics.
\end{enumerate}

Taken together, these results recast catastrophic forgetting not as an inevitable tax on specialization, but as a solvable optimization artefact—one that the model itself is surprisingly well equipped to fix. By letting an LLM “talk to itself’’ for a brief moment before adaptation, we unlock a simple yet effective path toward continually learning systems that remember \emph{and} advance.

\section{Related Work}
\paragraph{Synthetic Data.}

\begin{table*}[t]
\centering
\small
\setlength\tabcolsep{4pt}
\renewcommand{\arraystretch}{1.15}
\resizebox{0.85\textwidth}{!}{%
\begin{tabular}{cc*{2}{c}*{12}{c}}
\toprule
\multirow{4}{*}{Model} & \multirow{4}{*}{Regime} &
\multicolumn{2}{c}{\textbf{In-Domain}} &
\multicolumn{10}{c}{\textbf{General Benchmarks}} \\
\cmidrule(lr){3-4} \cmidrule(lr){5-14}
& & \multicolumn{2}{c}{Tasks (AR)} &
\multicolumn{2}{c}{GSM8K} & \multicolumn{2}{c}{MMLU} &
\multicolumn{2}{c}{IFEval} & \multicolumn{2}{c}{MedText} &
\multicolumn{2}{c}{AGIEval\_G} \\
\cmidrule(lr){3-4}
\cmidrule(lr){5-6} \cmidrule(lr){7-8}
\cmidrule(lr){9-10} \cmidrule(lr){11-12}
\cmidrule(lr){13-14} 
& & full & LoRA & full & LoRA & full & LoRA & full & LoRA & full & LoRA & full & LoRA  \\
\midrule

\multirow{6}{*}{LLaMA3.2-3B-I.} 
& Base               & \multicolumn{2}{c}{13.0} & \multicolumn{2}{c}{64.7} & \multicolumn{2}{c}{59.8} & \multicolumn{2}{c}{67.6}  & \multicolumn{2}{c}{24.9} & \multicolumn{2}{c}{36.2}  \\
& Task-only     & 38.7 & 44.3 & 59.7 & 57.8 & 58.4 & 59.2 & 46.1 & 49.2 & 23.2 & 23.8 & 29.3 & 29.3\\
\arrayrulecolor{gray!50}\cmidrule(lr){2-14}
\arrayrulecolor{black}
& Freeze   &44.1  &-  &  61.5 &  - &  59.5 & - &  43.8 &  - &24.4   &  - &28.9   & -  \\ 
& Task+Alpaca   &44.6  &44.5  &  56.2 &  56.9 &  58.9 & 58.4  &  59.8 &  53.3 &25.2   &25.3   &28.5   &29.7  \\ 
& Task+UC        & 44.5 & 44.5 &  62.2 &  62.5 &  \textbf{59.4} &  59.1 &  61.1 &  59.3 &  25.1 &  24.7 &  32.9 &  30.4 \\
& \methodname   & \textbf{45.6} & \textbf{44.7} &  \textbf{65.5} &  \textbf{65.6} &  59.3 &  \textbf{59.2} &  \textbf{63.2} &  \textbf{65.3} &  \textbf{25.2} &  \textbf{25.4} &  \textbf{34.6} &  \textbf{35.1}\\ \midrule
\multirow{6}{*}{LLaMA2-7B-C.} 
& Base              & \multicolumn{2}{c}{10.0}   & \multicolumn{2}{c}{23.0} & \multicolumn{2}{c}{47.2} & \multicolumn{2}{c}{44.6} & \multicolumn{2}{c}{24.6} & \multicolumn{2}{c}{24.3}  \\
& Task-only     & 50.4 & \textbf{51.8}& 10.9 & 16.9 & 46.7 & 46.6 & 0.7 & 3.6 & 22.7 & 24.0 & 22.0 & 22.3  \\
\arrayrulecolor{gray!50}\cmidrule(lr){2-14}
\arrayrulecolor{black}
& Freeze   &51.6  & - &  17.8 & - &  47.2 &  - &  1.5 &  - &24.6  & - &23.6 & -\\ 
& Task+Alpaca   &49.8  &48.3  & 15.8 &  16.4 &  45.8 &45.7   &  22.9 &  8.0 &24.8  &\textbf{25.0}  &23.6  &23.9 \\ 
& Task+UC       & 51.3 & 51.3&  \textbf{21.9} &  20.9 &  47.3 &  46.3 & \textbf{26.8} & 28.5 &  24.5 &  24.9 &  23.2 &  23.2  \\
& \methodname  & \textbf{52.2} & 51.7 &  21.3 &  \textbf{22.1} &  \textbf{47.9} &  \textbf{47.5} & 25.1 & \textbf{28.7} &  \textbf{24.9} &  24.9 &  \textbf{25.6} &  \textbf{28.5}  \\ \midrule
\multirow{6}{*}{Qwen2.5-7B-I.} 
& Base              & \multicolumn{2}{c}{20.2}  & \multicolumn{2}{c}{82.0} & \multicolumn{2}{c}{74.3} & \multicolumn{2}{c}{74.1} & \multicolumn{2}{c}{25.8} & \multicolumn{2}{c}{68.7}  \\
& Task-only     & 52.8 & 53.2  & 74.6 & 77.7 & 73.8 & 73.9 & 53.9 & 58.2 & 23.9 & 25.0 & 65.0 & 63.4  \\
\arrayrulecolor{gray!50}\cmidrule(lr){2-14}
\arrayrulecolor{black}
& Freeze   &30.7  & - & 57.7 & - &  73.6 & - & 47.7 & - &17.3  &  - &63.8 & -\\ 
& Task+Alpaca   &52.8  &53.3  & 72.6 &76.2  & 73.3 &73.4  &  56.8 &  62.6 &26.7   &26.8   &69.5   &66.7  \\ 
& Task+UC       & 52.7 & \textbf{53.4} &  73.0 &  76.1 &  73.6 &  \textbf{74.2} &  59.9 &  60.2 &  26.3 &  26.4 &  71.1 &  67.1  \\
& \methodname   & \textbf{52.9} & 49.8  &  \textbf{82.5} &  \textbf{82.1} &  \textbf{74.4} &  74.2 &  \textbf{68.5} &  \textbf{73.6} &  \textbf{26.8} &  \textbf{27.0} &  \textbf{72.0} &  \textbf{68.2}  \\ \midrule
\multirow{6}{*}{LLaMA3-8B-I.} 
& Base               & \multicolumn{2}{c}{12.2}  & \multicolumn{2}{c}{75.8} & \multicolumn{2}{c}{65.6} & \multicolumn{2}{c}{61.1} & \multicolumn{2}{c}{26.3} & \multicolumn{2}{c}{39.0} \\
& Task-only      &50.5 & 46.3  & 51.5 & 65.6 & 62.9 & 64.7 & 38.0 & 46.2 & 21.9 & 21.8 & 19.9 & 25.2   \\
\arrayrulecolor{gray!50}\cmidrule(lr){2-14}
\arrayrulecolor{black}
& Freeze   &48.7  & - &  41.1 & - &  63.3 & - &  41.1 & - &16.4   & -  &17.1   & - \\ 
& Task+Alpaca   & 43.9 &44.5  &  58.4 &  55.3 &  62.7 &63.6   &  58.4 &  55.3 &24.9   &24.7  &27.6   &23.6  \\ 
& Task+UC      &50.3 & 45.9   &  67.4 &  70.1 &  64.2 &  65.3 &  54.0 &  \textbf{62.7} &  24.9 &  24.5 &  \textbf{30.0} &  26.4  \\
& \methodname  &\textbf{50.8} & \textbf{49.5}  &  \textbf{68.2} &  \textbf{74.2} &  \textbf{64.7} &  \textbf{65.3} &  \textbf{63.2} &  57.6 &  \textbf{25.6} &  \textbf{26.1} &  29.8 & \textbf{29.7}  \\ \midrule
\multirow{6}{*}{LLaMA2-13B-C.} 
& Base              & \multicolumn{2}{c}{9.1}  & \multicolumn{2}{c}{35.7} & \multicolumn{2}{c}{53.6} & \multicolumn{2}{c}{46.0} & \multicolumn{2}{c}{25.8} & \multicolumn{2}{c}{26.4} \\
& Task-only      & 52.1 & 52.9 & 23.3 & 28.1 & 53.4 & 52.7 & 2.2 & 4.7 &24.6 &25.1 & 23.9 & 26.0 \\
\arrayrulecolor{gray!50}\cmidrule(lr){2-14}
\arrayrulecolor{black}
& Freeze   &52.2  & - &  26.6 &  - &  53.0 &  - &  1.2 &  - &25.2   &-   &\textbf{29.7}   &  -\\ 
& Task+Alpaca   &50.4  &52.3  &  25.7 &  27.1 &  52.7 & 52.2   &  21.5 &  12.0 &25.5  &25.3   & 28.0  &26.2  \\ 
& Task+UC       & 52.6 & 51.4 &  33.2 &  32.1 &  \textbf{53.9} &  \textbf{53.4} &  15.7 &  28.5 &  25.3 &  25.2 &  26.1 &  26.0 \\
& \methodname   & \textbf{53.3} & \textbf{52.9} &  \textbf{35.1} &  \textbf{35.3} &  53.7 &  53.3 &  \textbf{24.8} &  \textbf{37.1} &  \textbf{25.7} & \textbf{25.9} &  25.6  &  \textbf{26.4} \\ 

\bottomrule
\end{tabular}}
\caption{Performance of the three fine-tuning regimes on the in-domain test set and five general benchmarks. All general-ability scores are reported in the five-shot setting; accuracy is used for every benchmark except MedText, where ROUGE-L is reported. Within each task block, the best result among the five regimes is \textbf{bold}.}
\label{tab:main}
\end{table*}

Leveraging synthetic data has emerged as an effective strategy to improve and adapt large language models without extensive human annotations. For example, \cite{selfinstruct} introduce Self-Instruct, a framework where a model generates its own instruction-following data to bootstrap further fine-tuning. Magpie \cite{DBLP:conf/iclr/XuJNDP0L25} introduces a self-synthesis method to generate large-scale alignment data by utilizing only pre-defined chat templates. Crescent \cite{DBLP:journals/corr/abs-2502-13441} proves that self-generated data has the potential to enable the model to self-improve on reasoning tasks such as mathematics. Such synthetic data generation techniques provide a nearly annotation-free approach to align pretrained models with desired behaviors, laying a foundation for our approach of model-in-the-loop data synthesis.

\paragraph{Mitigation Methods of Catastrophic Forgetting.}

Catastrophic forgetting is a well-known challenge in fine-tuning LLMs on narrow domains, where pre-trained general knowledge and capabilities can substantially degrade \cite{song2025alleviate}. A common mitigation strategy is data rehearsal, which mixes general-domain or instruction-tuning data with task-specific data during fine-tuning to preserve prior abilities. This idea has been explored through both external and self-generated rehearsal data. For example, Self-Synthesized Rehearsal \cite{DBLP:conf/acl/HuangCWYLSYS24} generates auxiliary samples to reduce forgetting, though it still relies on access to high-quality reference data from earlier stages. In practice, many domain-specific LLMs similarly replay portions of general instruction data during adaptation, a pattern observed across legal, financial, and other applied settings \cite{sdu_fuzi_mingcha,keexpanding,DBLP:journals/corr/abs-2506-09428}.

Beyond data-centric approaches, parameter-based methods aim to constrain model updates during fine-tuning. Inspired by continual learning, some methods regularize parameter drift to protect important weights, such as EWCLoRA \cite{DBLP:conf/nips/XiangTGSWYH23}. Others rely on parameter-efficient fine-tuning schemes, including adapters or partial weight updates, to inherently retain pre-trained knowledge by limiting the number of trainable parameters \cite{DBLP:conf/iclr/HuSWALWWC22}.

\section{Methodology}
\label{sec:sasft}

\begin{figure*}[t]
\centering
\begin{subfigure}[t]{0.47\linewidth}
    \centering
    \includegraphics[width=0.8\linewidth]{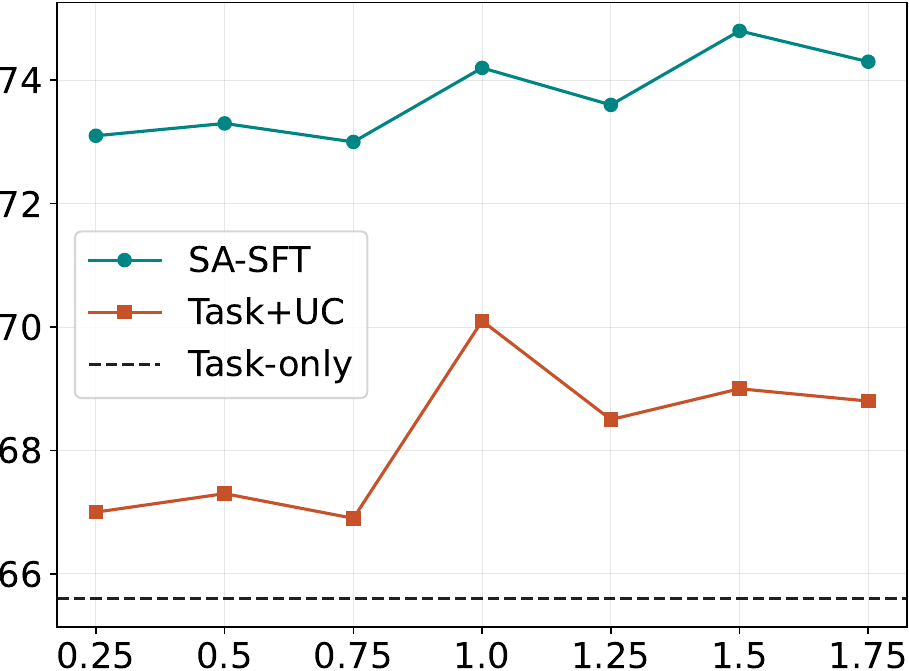}
    \caption{GSM8K accuracy (\%).}
    \label{fig:ratio-left}
\end{subfigure}
\hfill
\begin{subfigure}[t]{0.47\linewidth}
    \centering
    \includegraphics[width=0.8\linewidth]{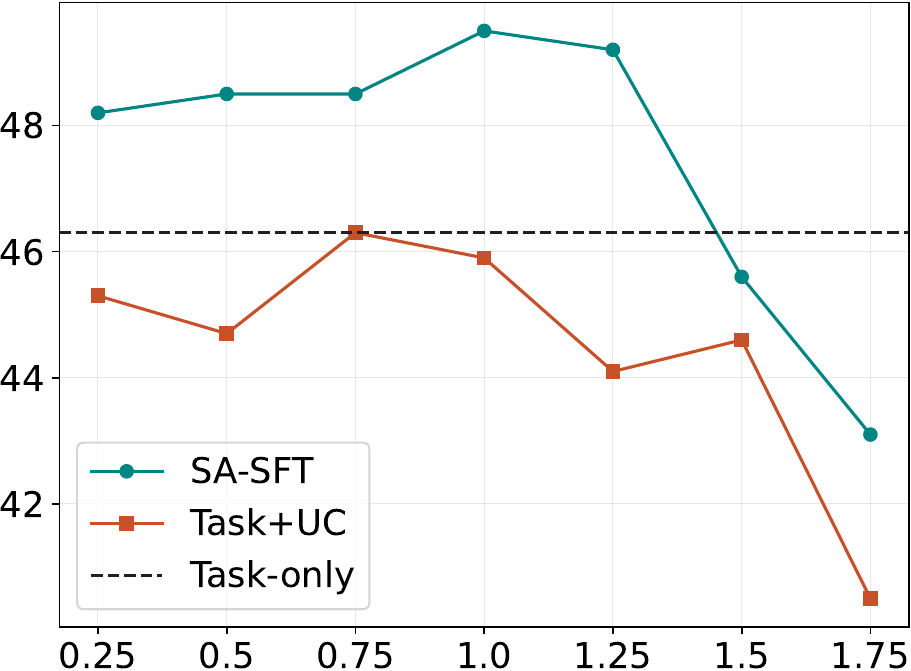}
    \caption{In-domain results (AR).}
    \label{fig:ratio-right}
\end{subfigure}
\caption{Results of \methodname and Task+UC under varying mix ratios. The dashed line indicates the Task-only baseline.}
\label{fig:ratio-dual}
\end{figure*}

We refer to our proposed procedure as \emph{Self-Augmented Supervised Fine-Tuning} (\textbf{\methodname}). {\methodname} inserts a single \emph{self-augmentation} phase before conventional supervised fine-tuning (SFT). The model first fabricates a small \emph{refresher} dialogue corpus from its own knowledge, then this corpus is \emph{mixed} with the downstream data. The resulting dataset, requiring no additional post-processing, is then directly fed into the model for fine-tuning.

\paragraph{Notation.}
We denote by $f_{\omega_0}$ a pretrained language model with parameters $\omega_0$,
by $\mathcal{D}_{\text{task}}=\{(x_i,y_i)\}_{i=1}^{N}$ the domain-specific SFT dataset,
and by $\lambda\in(0,+\infty]$ the \emph{mix-in ratio}.

\paragraph{Step 1: Self-augmentation.}
Sample $M=\lambda N$ question–answer pairs with the frozen model:
\begin{equation}
\left\{
\begin{aligned}
x'_j &\sim f_{\omega_0}\!\bigl(\,\cdot \mid p_j\bigr) \\
y'_j &\sim f_{\omega_0}\!\bigl(\,\cdot \mid x'_j\bigr)
\end{aligned}
\right.
\quad j = 1,\dots,M,
\end{equation}
where each prompt $p_j$—either a Magpie \cite{DBLP:conf/iclr/XuJNDP0L25} special token or a Crescent \cite{DBLP:journals/corr/abs-2502-13441} bait prompt—elicits a continuation treated as $x'_j$.  Feeding $x'_j$ back once produces the answer $y'_j$.  We denote the resulting corpus $\mathcal{D}_{\text{self}}= \{(x'_j,\,y'_j)\}_{j=1}^{M}$.

\paragraph{Generation hyper-parameters.}
Unless stated otherwise, all self-generated samples use nucleus (top-$p$) sampling with $p{=}0.9$, temperature $T{=}0.7$, and a maximum length of 256 tokens; we found these default values balance diversity and faithfulness across all base models tested.

\paragraph{Step 2: Mix-in.}
Merge the two corpora
\begin{equation}
  \mathcal{D}_{\text{mix}}
  \;=\;
  \mathcal{D}_{\text{task}}\;\cup\;\mathcal{D}_{\text{self}},
  \lvert\mathcal{D}_{\text{mix}}\rvert
  \;=\;
  N(1+\lambda).
\end{equation}

\paragraph{Fine-tuning objective.}
Optimise parameters (full or LoRA) under the usual cross-entropy

\begin{equation}
\begin{aligned}
\min_{\omega}\quad
&\underbrace{\frac{1}{(1+\lambda)N}\!\sum_{(x,y)\in\mathcal{D}_{\text{task}}}
\!\ell(f_{\omega}(x),y)}_{\text{task loss}} \\
\;+\;
&\underbrace{\frac{1}{(1+\lambda)N}\!\sum_{(x,y)\in\mathcal{D}_{\text{self}}}
\!\ell(f_{\omega}(x),y)}_{\text{self-augmentation loss}}.
\end{aligned}
\end{equation}

The objective corresponds to standard SFT on the concatenation of $\mathcal D_{\text{task}}$ and $\mathcal D_{\text{self}}$, i.e., minimizing the average token-level cross-entropy over the union of the two datasets. The relative contribution of each component is implicitly determined by its dataset size, with $|\mathcal D_{\text{task}}|=N $ and $|\mathcal D_{\text{self}}|=\lambda N$.

\paragraph{Implementation notes.}
{\methodname} is model-agnostic and can be applied atop any pretrained decoder-only architecture without modifying the base model or training framework. The self-augmentation phase requires no gradient updates and can be implemented as a lightweight sampling loop. Moreover, the ratio $\lambda$ serves as a simple yet flexible knob to control the influence of self-generated data, allowing practitioners to adapt {\methodname} to a wide range of domains and data regimes with minimal tuning effort.

\section{Main Experiments}
\label{sec:mainexp}

\subsection{Experimental Setup}

\paragraph{Models.}

\begin{table}[t]
\centering
\small
\setlength\tabcolsep{5pt}
\renewcommand{\arraystretch}{1.15}
\resizebox{\linewidth}{!}{
\begin{tabular}{ccccccc}
\toprule
Generator Model  & In-Domain & GSM8K & MMLU & IFEval & MedText & AGIEval\_G \\
\midrule
LLaMA3-8B-Instruct             & \underline{49.5} & \textbf{74.2} & \textbf{65.3} & 57.6 & \textbf{26.1} & \textbf{29.7} \\
\arrayrulecolor{gray!90}\cmidrule(lr){1-7}
\arrayrulecolor{black}
Qwen2.5-7B-Instruct             & 44.6 & 62.4 & 62.5 & \underline{58.0} & 25.7 & 24.8 \\
LLaMA2-7B-Chat       & 49.5 & 50.3 & 60.3 & 54.3 & 25.9 & 25.6 \\
LLaMA2-13B-Chat      & \textbf{50.8} & 56.8 & 61.0 & 53.6 & 25.9 & 26.4 \\
LLaMA3.2-3B-Instruct  & 49.1 & 65.7 & 60.8 & 55.9 & 26.0 & 26.4 \\
LLaMA3-70B-Instruct  & 46.9 & \underline{72.8} & \underline{63.1} & \textbf{60.8} & \underline{26.0} & \underline{29.2} \\
\bottomrule
\end{tabular}
}
\caption{
\textbf{Impact of generator choice on downstream performance.}
All settings use LLaMA3-8B-Instruct as the target model, fine-tuned with LoRA on a downstream task mixed 1:1 with synthetic dialogues from the generator model.
We report in-domain accuracy and generalization performance on five benchmarks (all 5-shot).
Best generalization results are \textbf{bolded}.
}
\label{tab:generator-impact}
\end{table}

We evaluate five mainstream instruction-tuned LLMs that cover two generations and three parameter scales: LLaMA3-8B-Instruct, LLaMA3.2-3B-Instruct \cite{DBLP:journals/corr/abs-2407-21783}, Qwen2.5-7B-Chat \cite{DBLP:journals/corr/abs-2412-15115}, LLaMA2-7B-Chat, and LLaMA2-13B-Chat \cite{DBLP:journals/corr/abs-2307-09288}. All checkpoints are taken from their original open-source releases and are kept frozen for the “Base’’ baseline.


\paragraph{Fine-tuning tasks.}
We sample five heterogeneous tasks from the Super-NaturalInstructions (SuperNI) benchmark \cite{DBLP:conf/emnlp/WangMAKMNADASPK22} as our fine-tuning suite. SuperNI is a large-scale collection of 1,616 diverse NLP tasks, each accompanied by expert-written instructions, designed to evaluate a model’s ability to generalize across task boundaries. The benchmark spans 76 distinct task types, including classification, extraction, sequence tagging, infilling, rewriting, and more, making it a rigorous testbed for instruction-following capabilities. Among these, we select five representative tasks that cover a broad functional spectrum:
\begin{enumerate*}

    \item Question Answering (QA); 
    \item Question Generation (QG);
    \item Sentiment Analysis (SA);
    \item Summarization (Sum.);
    \item Translation (Trans.);
\end{enumerate*}

For training, we randomly sample 2,000 instruction response pairs per task (10,000 in total), and reserve an additional 500 pairs per task (2,500 total) for in-domain evaluation. These tasks were deliberately chosen to reflect varied linguistic phenomena and task structures, thereby simulating realistic multi-task fine-tuning scenarios.

\paragraph{Fine-tuning regimes.}
For every model we train two adaptation flavours: full-parameter supervised fine-tuning (full) and parameter-efficient Low-Rank Adaptation (LoRA) \cite{DBLP:conf/iclr/HuSWALWWC22} with rank 8 and $\alpha$ = 16. Within each flavour we consider six data regimes:

\begin{itemize*}
    \item \textbf{Base:} no additional training;
    \item \textbf{Task-only:} fine-tune on the task corpus alone;
    \item \textbf{Freeze:} identical to the \textit{Task-only} setting but with the bottom 6 layers, including the embedding layer, frozen during training, following the layer-freezing settings proposed in prior work \cite{DBLP:conf/iclr/ZhengCQ025}, and applied only to \emph{full} SFT.
    \item \textbf{Task+Alpaca:} mix the task corpus 1:1 with the \textsc{Alpaca} instruction-following dataset (20,000 pairs in total);
    \item \textbf{Task+UC:} mix the task corpus 1:1 with \textsc{UltraChat} (20000 pairs in total);
    \item \textbf{\methodname:} mix the task corpus 1:1 with the $\mathcal{D}_{\text{self}}$ from \methodname procedure (20000 in total).
\end{itemize*}

We use the LLaMA-Factory \cite{zheng2024llamafactory} framework for all fine-tuning experiments.
Training and evaluation are all performed on NVIDIA A100-80 GB GPUs.
All models are trained for 3 epochs with a batch size of 32 under Deepspeed ZeRO-3 \cite{DBLP:conf/kdd/RasleyRRH20}  and a cosine schedule with 10 \% warm-up; the learning rate is $2 \times 10^{-4}$ for LoRA and $1 \times 10^{-5}$ for full SFT.
All computations use \texttt{bfloat16} precision.

\paragraph{Generation Configurations.}

For the LLaMA3 family (8B and 3.2B) and Qwen2.5, we generate synthetic data with the Magpie sampling scheme, adopting the same hyper-parameters released in its official repository \cite{DBLP:conf/iclr/XuJNDP0L25}. Magpie is not fully compatible with the LLaMA2 series, so for LLaMA2-7B and LLaMA2-13B we switch to the \textsc{Crescent}. We set the bait prompt as: \enquote{\emph{Give me some questions about everyday life}}, while all remaining settings follow the values reported in the \textsc{Crescent} paper \cite{DBLP:journals/corr/abs-2502-13441}.

\paragraph{Benchmarks.} 
Our evaluation comprises two parts. First, to confirm in-domain effectiveness, we test every model produced by each fine-tuning regime on the in-domain test splits of the five SuperNI tasks and report their \emph{Averaged ROUGE-L} (\textbf{AR}) score following the evaluation settings in \cite{DBLP:conf/acl/HuangCWYLSYS24}. Second, to quantify catastrophic forgetting, we compare all the models on a suite of broad benchmarks—GSM8K \cite{cobbe2021training}, MMLU \cite{hendryckstest2021}, IFEval \cite{zhou2023instructionfollowing}, MedText \cite{melamud2019automaticgenerationshareablesynthetic}, and the \emph{Gaokao Chinese} subset of AGIEval (AGIEval\_G) \cite{zhong2023agieval}—covering mathematical reasoning, multi-domain knowledge, instruction following and other general-purpose knowledge and abilities, where we report accuracy for all benchmarks except MedText, for which we report ROUGE-L.

\paragraph{Results and Discussion}

As shown in \cref{tab:main}, our results fall into two complementary views: \textbf{in-domain efficacy} and \textbf{general-ability retention}.

\textbf{In-domain efficacy.} Across the ten in-domain settings (five models $\times$ full \& LoRA), all five fine-tuning regimes—Task-only, Freeze, Task + Alpaca, Task + UC, and \methodname outperform the untouched bases. 
Every fine-tuning variant surpasses the Base models by a large margin. The \textbf{Freeze} variant achieves comparable yet consistently lower scores than Task-only, indicating that restricting lower-layer adaptation hampers the model’s ability to efficiently acquire new task knowledge.

Meanwhile, introducing additional conversational data produces markedly different effects depending on its origin. \textbf{Task+Alpaca} and \textbf{Task+UC} achieve performance comparable to \textbf{Task-only} across most settings. However, both remain slightly below the pure \textbf{Task-only} regime on average, suggesting that while cross-domain data helps maintain learning capacity, it introduces mild domain interference that limits full specialization. Most importantly, \textbf{\methodname} continues to outperform all other regimes in \textbf{8 out of 10} in-domain configurations, confirming that self-generated data not only preserves domain focus but also accelerates targeted knowledge acquisition—echoing findings from SCAR \cite{li2024scar}.

\textbf{General-ability retention.}
Turning to the general benchmarks, we evaluate 50 different scenarios (5 models $\times$ 5 benchmarks $\times$ 2 adaptation flavours). \textbf{Task-only} fine-tuning consistently degrades performance relative to Base (Task-only $<$ Base), reaffirming the ubiquity of catastrophic forgetting. \textbf{Freeze} provides an unstable form of forgetting mitigation. While freezing the bottom six layers can, in some cases, reduce representational drift and preserve parts of general ability, its effect is highly sensitive to the base model. For instance, on Qwen2.5-7B-I., the \textbf{Freeze} variant consistently underperforms \textbf{Task-only} across all benchmarks, and even suffers abrupt collapses on specific tasks such as \textsc{GSM8K}. These observations indicate that layer freezing offers only limited and inconsistent protection against forgetting.
On the other hand, mixing task data with existing instruction or dialogue corpora proves more effective overall. Both \textbf{Task+Alpaca} and \textbf{Task+UC} alleviate forgetting to varying degrees, but their success depends heavily on the alignment between auxiliary and target domains. For example, on MedText, mixing \textbf{Alpaca} surpasses \textbf{UltraChat} in all 10 cases, whereas on the mathematical \textsc{GSM8K}, \textbf{UltraChat} outperforms \textbf{Alpaca} in 9 out of 10 cases. This illustrates a key limitation of conventional data-mixing strategies: while capable of mitigating forgetting, they demand \emph{careful dataset selection and tuning to remain effective across domains}.

In contrast, \textbf{\methodname} robustly mitigates forgetting in \textbf{all 50 / 50} cases and achieves the best overall performance in \textbf{40 / 50} scenarios, maintaining or improving general-ability metrics across \textbf{model scale} (3.2B–13B), \textbf{architecture} (LLaMA2, LLaMA3, Qwen2.5), and \textbf{evaluation domain}. Its effectiveness remains consistent across different base models and benchmark types, showing no pronounced sensitivity to model size, backbone design, or task nature. This stability suggests that the self-generated data from \methodname\ captures transferable knowledge that regularizes fine-tuning in a model-agnostic manner, offering a universal and reliable defence against catastrophic forgetting without requiring dataset-specific tuning.


Together, these results demonstrate that letting a model self-augment via a \emph{task-aware}, \emph{lightly curated} synthetic corpus is a more effective and self-consistent \textbf{defence against catastrophic forgetting} than external data mixing or layer freezing, while still \emph{enhancing in-domain performance}.

\section{Analysis}

\subsection{Sensitivity to the \methodname Mix Ratio}


To assess how sensitive \methodname is to the amount of self-generated data and whether \methodname's benefit over UltraChat emerges only at a narrow span, we carry out a mix-ratio sensitivity study. 

Starting from the task-specific training set $D_{\text{task}}$, we create two augmented variants: (1) \textbf{Task+UC:} $D_{\text{task}}$ mixed $1 : r$ with UltraChat data; (2) \textbf{\methodname:} $D_{\text{task}}$ mixed $1 : r$ with self-generated data;
where the ratio $r\in \{0.25\allowbreak,0.50\allowbreak,0.75\allowbreak,1.00\allowbreak,1.25\allowbreak,1.50\allowbreak,1.75\allowbreak\}.$
For every setting we fine-tune LLaMA3-8B-I. with LoRA under settings in \cref{sec:mainexp}, yielding 14 checkpoints (7 ratios × 2 sources). We then measure GSM8K and in-domain performance on those models. The results are showed in \cref{fig:ratio-dual}.

\paragraph{Results.}
Across both panels, we have the following key findings:
\begin{enumerate}
    \item \methodname consistently outperforms the Task+UC at every tested ratio, indicating that its advantage is \emph{not confined to a narrow sweet-spot}.
    \item \cref{fig:ratio-left} shows a general rise in GSM8K accuracy for both mixing strategies as the ratio $r$ increases: \emph{adding more generic data can better restore general capabilities}.
    \item \cref{fig:ratio-right} shows a sharp drop once $r>1.25$. Excessive generic data begins to \emph{dilute task-specific signal}, harming downstream tasks.
\end{enumerate}

Balancing these observations, a 1 : 1 mix ($r=1.0$) offers the best trade-off—\emph{maximizing general-domain retention without eroding in-domain accuracy}. Crucially, \methodname’s self-generation feature grants practitioners fine-grained control over the data size, allowing them to tailor the mix to any task \emph{without hunting for external corpus}.

\subsection{Step-Level Trajectory of Forgetting}

\begin{figure}[t]
  \centering
  \includegraphics[width=\linewidth]{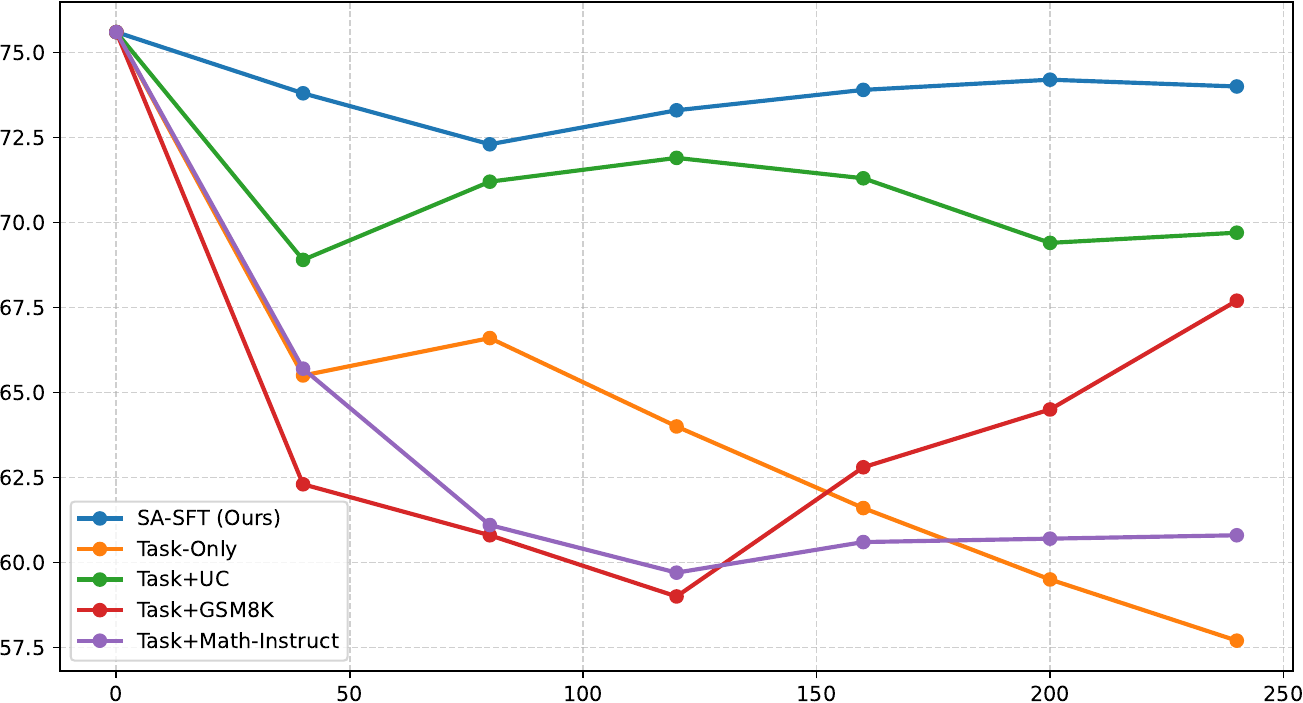}
  \caption{GSM8K accuracy over training steps.}
  \label{fig:time_course}
\end{figure}

To pinpoint \emph{when} catastrophic forgetting emerges and \emph{how} different fine-tuning regimes alter its course, we fine-tuned LLaMA3-8B-Instruct for three epochs and saved a snapshot every 40 optimisation steps (0 $\rightarrow$ 240). At each snapshot we evaluated GSM8K accuracy. The experiment compared five otherwise identical regimes that differ only in the auxiliary data blended with the downstream task corpus: (\romannumeral 1)\textbf{Task-only}; (\romannumeral 2) \textbf{Task+UC} (1:1 mix); (\romannumeral 3) \textbf{\methodname}-our method (1:1 mix); (\romannumeral 4) \textbf{Task+Math-Instruct} \cite{DBLP:conf/iclr/YueQZFH00C24} (1:1 mix of an external mathematics corpus); (\romannumeral 5) \textbf{Task+GSM8K} (approximately 1:0.7 mix of GSM8K’s own train split);

\cref{fig:time_course} yields the following insights:
\begin{enumerate*}
    \item The \methodname curve remains \emph{essentially flat and consistently achieves the highest accuracy}, indicating that self-generated data can provide stable and robust protection for math reasoning throughout the \textbf{entire} fine-tuning process.
    \item The Task+UC curve \emph{consistently under performs} compared to \methodname and exhibits \emph{greater fluctuation}, suggesting \textbf{less stable resistance} to forgetting.
    \item Task-only exhibits the canonical monotonic decline, confirming significant forgetting when fine-tuning on specific downstream tasks without any augmentation.
    \item Task+GSM8K follows a U-shaped trajectory: performance drops sharply at first and then gradually recovers, indicating that mixing data from \emph{even the same distribution} as the evaluation set does not prevent forgetting, but instead causes the model to forget and then learn new knowledge (or adapt to new distribution). This recovery process is slow, in our experiment, even by step 240—when the model has already fit the fine-tuning task well—\emph{it still has not fully regained its math ability}.
    \item Task+Math-Instruct improves \textbf{marginally over Task-Only}, implying that simply injecting high-quality—but \emph{distribution-mismatched}—math corpus cannot restore forgotten skills within the 240-step budget; much of the compute is likely spent adapting to \emph{new style}, rather than reinforcing \emph{transferable reasoning skills}.
\end{enumerate*}

\paragraph{Implications.}
The contrast between \methodname’s flat trajectory and the recovery-after-collapse seen with Task+GSM8K highlights a crucial distinction: \emph{preventing forgetting during training versus repairing it afterwards}. Generating a model-specific corpus before fine-tuning offers a lightweight, computation-efficient safeguard that outperforms both open-domain and in-domain external datasets.

\subsection{Impact of Generator Model}

One of our core claim: \methodname is most effective when the synthetic corpus is produced by \textbf{the very model that will later be fine-tuned}. A natural follow-up question is: \emph{How sensitive is this benefit to the choice of generator?} If we depart from the \enquote{same-model} principle—using a generator that differs in architecture, training recipe, or parameter size—does the resulting synthetic data still curb forgetting, or does the mismatch degrade its value?

To answer this, we fix LLaMA3-8B-Instruct as the base model and consider five alternative generators:
(1) Qwen2.5-7B-I., (2) LLaMA2-7B-C., (3) LLaMA2-13B-C., (4) LLaMA3.2-3B-I., (5) LLaMA3-70B-I..
\begin{table}[t]
\centering
\resizebox{\linewidth}{!}{
\begin{tabular}{@{}ccccc@{}}
\toprule
\textbf{Model} & \textsc{Math} & \textsc{Logic} & \textsc{Code} &
\textsc{Other}\\
\midrule
LLaMA3-8B-I.
  &  45  & 17 & 38 & 1900\\
Qwen2.5-7B-I.
  & 74 & 29  & 118  & 1779\\
\bottomrule
\end{tabular}
}
\caption{Density of math/logic/code content in
\(2\,000\) self-generated dialogues (absolute counts)}
\label{tab:logic-density}
\end{table}
For every generator we keep all other hyper-parameters identical to the \cref{sec:mainexp}; the synthetic corpus size is matched 1 : 1 to the fine-tuning tasks. Each combined dataset is then fine-tuned on the base model using exactly the same LoRA configuration as before. Finally, we evaluate: (i) in-domain performance on the task datasets and (ii) general competence on GSM8K, MMLU, IFEval, MedText, and AGIEval\_G (5-shot, MedText reported with ROUGE-L).

\cref{tab:generator-impact} corroborates our central claim. Across the five general benchmarks, the strongest protection against forgetting (four of the five best scores) arises when the synthetic corpus is produced by the same LLaMA3-8B-Instruct model that is later fine-tuned. The next-best result comes from data generated by the much larger LLaMA3-70B-Instruct, yet still falls short of the self-generated variant.

On the in-domain task, self-augmentation again performs competitively (second only to the LLaMA2-13B-C. generator), while the latter suffers a severe drop on the general benchmarks—a trade-off that self-generated data avoids. It is to be noted that, on public leaderboards \cite{open-llm-leaderboard-v2} , Qwen2.5-7B-I. and LLaMA3-70B-I. both outrank the 8B model. Their synthetic data indeed sharpen instruction-following ability (see the IFEval column), but they do not replicate the \textbf{unique anti-forgetting effect} of \enquote{same-model} data. 

Taken together, these results indicate that higher-quality but stylistically mismatched data cannot match the effectiveness of truly self-generated data.

\subsection{What Really Shields Models from Forgetting?}
\label{sec:shield}
The preceding experiments show that inserting a short \methodname{} phase
almost completely restores GSM8K accuracy, but \emph{why} does it work?
Two competing explanations exist:
\begin{enumerate}[itemsep=8.5pt, topsep=8.5pt]
    \item[A.] The self-generated corpus happens to contain \textbf{critical
    facts and skills} (e.g.\ mathematical reasoning) that are later
    replayed during SFT, directly inoculating the model.
    \item[B.] The corpus merely matches the model’s own \textbf{latent
    distribution}; any low–perplexity text—regardless of topic—tempers
    the large, style–driven parameter shifts responsible for
    \emph{format-induced forgetting}.
\end{enumerate}
We distinguish A and B via a two-step study:

\paragraph{Step 1: How Much Math/Logic Is in the Data?}

\paragraph{Protocol.}
We select tow generators: \textbf{LLaMA3-8B-I.} and
\textbf{Qwen2.5-7B-I.}—we uniformly
sample \(2\,000\) dialogues.
Each dialogue is classified by GPT-4o, which assigns one of four mutually
exclusive labels: (\romannumeral 1)\textsc{Math} (arithmetic, algebra, word problems); (\romannumeral 2) \textsc{Logic} (syllogisms, truth tables, grid puzzles); (\romannumeral 3)\textsc{Code} (programming, debugging, algorithms); (\romannumeral 4)\textsc{Other} (all remaining content).

\paragraph{Findings.}
\cref{tab:logic-density} reveals that strictly
\emph{math-related} material is scarce:
only \(\approx5\,\%\) of the self-generated dialogues touch
mathematics, logic, or code for LLaMA3, (\(\approx11\,\%\) for Qwen2.5).
Despite this modest share, \cref{sec:mainexp} showed a great recovery on GSM8K, hinting that factors beyond mere replay of math problems may be at work.

\paragraph{Step 2: Removing \emph{All} Math/Logic/Code}

\paragraph{Construction of “logic-free’’ corpora.}
Using the same GPT-4o filter, we \emph{delete} every dialogue labelled \textsc{Math}, \textsc{Logic}, or \textsc{Code} from the self-generated corpora of these two models.
We then down-sample the data to a fixed
\(10\,000\) pairs to keep the ratio to 1:1,
obtaining two new datasets:
\textbf{LLaMA3-NoLogic} and \textbf{Qwen2.5-NoLogic}.

\paragraph{Fine-tuning.}
We repeat LoRA fine-tuning under the identical settings of
\cref{sec:mainexp} and evaluate on GSM8K on LLaMA3-8B-I. and Qwen2.5-7B-I. models.
\cref{tab:gsm8k-nologic} shows the outcome.

\begin{table}[t]
\centering
\resizebox{\linewidth}{!}{
\begin{tabular}{@{}lcc@{}}
\toprule
\textbf{Regime} & LLaMA3-8B-I. & Qwen2.5-7B-I.\\
\midrule
Base &75.8 &82.0\\
\arrayrulecolor{gray!50}\cmidrule(lr){1-3}
\arrayrulecolor{black}
Task-Only                         & 65.6 & 77.7\\
\methodname \textsc{Full data}              & 74.2 & 82.1\\
\methodname \textsc{NoLogic}           & 73.7 & 82.3\\
\bottomrule
\end{tabular}
}
\caption{GSM8K accuracy (\%) after SFT with and without
math/logic/code content.}
\label{tab:gsm8k-nologic}
\end{table}

\textbf{Key result:} Even after purging every trace of
math/logic/code, \textbf{\methodname data still prevents most of the GSM8K accuracy drop}.

We therefore support explanation B: we argue that a substantial portion of catastrophic forgetting in SFT arises from \emph{style-induced gradient drift} rather than the loss of underlying knowledge. By mixing self-generated samples that are pretraining-aligned in surface form, \methodname counteracts such spurious stylistic updates, thereby preserving general capabilities while learning the target task. Theoretical analysis is provided in \cref{sec:theory}.







\section{Conclusion}
Catastrophic forgetting need not be the price of specialization. With a brief “self-talk’’ phase, our \textbf{Self-Augmented SFT} (\methodname) shields LLMs from skill loss while boosting task accuracy—without external data, extra losses, or manual curation. Across five public benchmarks, \methodname not only recovers general ability and strengthens in-domain performance, but also consistently outperforms mainstream baselines, offering the most stable balance between retention and adaptation. Our theoretical analysis further explains this robustness: self-generated, in-distribution text implicitly regularizes training by counteracting style-induced parameter drift. Because \methodname integrates seamlessly into existing fine-tuning pipelines with minimal overhead, it provides an immediate, low-effort safeguard for building stable and adaptive LLMs.

\section{Limitations}
\label{sec:limitations}
While \methodname is simple and effective, it also has several limitations.

\paragraph{Dependence on the base model’s generative quality.}
\methodname relies on the frozen base model to produce a self-authored corpus that is sufficiently diverse and well-formed. For weaker or poorly instruction-tuned models, self-dialogues may become repetitive, low-information, or biased toward shallow conversational patterns, which can reduce the regularization effect and may even introduce noise into training. In such cases, \methodname may require more careful prompt design or basic filtering to avoid degenerate samples.


\paragraph{Incomplete coverage of all forgetting mechanisms.}
Our analysis highlights \emph{style-induced parameter drift} as an important contributor to catastrophic forgetting, and \methodname is particularly suited to counteracting this failure mode via self-aligned text. However, the \emph{true causes} of catastrophic forgetting are still actively debated and likely multifactorial: existing work attributes forgetting to different sources, including gradient interference and representation drift, task similarity and sequential training effects, and the interaction between optimization dynamics and parameter importance \cite{kemker2018measuring,ramasesh2021effect,DBLP:journals/corr/abs-2308-08747,DBLP:conf/iclr/ZhengCQ025}. As a result, \methodname should be viewed as a targeted mitigation for the mechanism we identify and formalize. While our empirical results suggest that self-aligned rehearsal can be effective in many practical settings, we do not claim it resolves all possible causes of forgetting, nor do we deeply investigate regimes where other mechanisms dominate.

\paragraph{Extra generation cost and potential data risks.}
Although \methodname avoids external datasets and additional loss terms, it still introduces an upfront sampling step to create $\mathcal{D}_{\text{self}}$, which incurs extra inference-time compute and storage—especially when scaling to many tasks or continual updates. Moreover, self-generated data can contain inaccuracies, unsafe content, or spurious biases inherited from the base model, which may be amplified when mixed into fine-tuning. Practical deployments may therefore benefit from inexpensive safeguards (e.g., deduplication, toxicity screening, or simple quality heuristics) to reduce such risks without undermining the method’s lightweight nature.


\appendix

\section{Theoretical Analysis: Style-Mismatch–Induced Forgetting}
\label{sec:theory}

\subsection{Intuitive Explanation}

When fine-tuned on external SFT datasets, a pretrained language model often encounters stylistic conventions—templated instruction formats, repeated punctuation, or specific discourse patterns—that rarely appear in its pretraining corpus. These stylistic artifacts trigger large gradient updates as the optimizer attempts to mimic the new “surface form.” Such updates, however, carry little semantic information and primarily distort parameter directions that were tuned to the pretraining style.
Consequently, the model experiences catastrophic forgetting: performance on general reasoning and pretraining-aligned tasks declines, not because the underlying knowledge is overwritten, but because the parameters have been rotated toward an unfamiliar stylistic regime.

Our framework \methodname proposes that self-generated mixing alleviates this issue. When we interleave a portion of self-generated samples—whose linguistic style matches the model’s own pretraining distribution—the optimizer’s updates become more style-balanced. Gradients arising from familiar, in-distribution text statistically cancel the superficial stylistic components introduced by the external SFT corpus. As a result, the parameter trajectory remains closer to the pretraining manifold, preserving previously acquired capabilities while still absorbing task-specific information.

\subsection{Formal Setup}

Let $f_{\omega}$ be a pretrained model with parameters $\omega_0$, and let $\ell(\cdot,\cdot)$ denote the per-token cross-entropy loss.
For a distribution $\mathcal{D}$,
\begin{equation}
\begin{aligned}
\mathcal{R}_{\mathcal{D}}(\omega) = \mathbb{E}_{(x,y)\sim \mathcal{D}}[\ell(f_{\omega}(x),y)].
\end{aligned}
\end{equation}
Denote by $\mathcal{D}_0$ the pretraining distribution and by $\mathcal{D}_{\text{task}}$ the external SFT dataset.
After $T$ fine-tuning steps from initialization $\omega_0$, the amount of forgetting is quantified as:
\begin{equation}
\begin{aligned}
\Delta_{\text{forget}}
= \mathcal{R}_{\mathcal{D}_0}(\omega_T) - \mathcal{R}_{\mathcal{D}_0}(\omega_0).
\end{aligned}
\end{equation}
The expected gradient at initialization is:
\begin{equation}
\begin{aligned}
g_{\text{task}}(\omega_0)
= \mathbb{E}_{(x,y)\sim \mathcal{D}_{\text{task}}}
[\nabla_{\omega}\ell(f_{\omega_0}(x),y)].
\end{aligned}
\end{equation}
We interpret each example as generated from latent \textbf{content} $c$ and \textbf{style} $s$, where $s$ captures superficial linguistic factors such as formatting or register.

\subsection{Style–Semantic Decomposition}

Empirically, gradients can be approximately decomposed into two subspaces:

\textbf{Assumption 1 (Approximate subspace separability):} There exists a \textbf{Style subspace} $\mathcal{S}\subset\mathbb{R}^d$ and its orthogonal complement $\mathcal{S}^{\perp}$ (the \textbf{Semantic} subspace) such that
\begin{equation}
\begin{aligned}
g_{\text{task}} \approx g_{\text{sty}} + g_{\text{sem}},
\quad
g_{\text{sty}}\in\mathcal{S};
g_{\text{sem}}\in\mathcal{S}^{\perp}.
\end{aligned}
\end{equation}
Here $g_{\text{sty}}$ captures gradients that primarily change surface form, while $g_{\text{sem}}$ modifies content-related representations.
When the stylistic distribution of $\mathcal{D}_{\text{task}}$ deviates strongly from the pretraining distribution $S_0$, the Style component $g_{\text{sty}}$ tends to dominate, producing parameter drift unrelated to semantic improvement.

To quantify this mismatch, we define the \textit{style perplexity gap}:

\begin{equation}
\begin{aligned}
\Delta_{\mathrm{ppl}} = {} &\mathbb{E}_{x\sim \mathcal{D}_{\text{task}}}[-\log p_{\omega_0}(x)]  \\
-&\mathbb{E}_{x\sim \mathcal{D}_0}[-\log p_{\omega_0}(x)].
\end{aligned}
\end{equation}

A larger $\Delta_{\mathrm{ppl}}$ indicates that the SFT text is less typical under the base model’s pretraining style, correlating with larger $||g_{\text{sty}}||$ and stronger forgetting.

\subsection{First-Order View of Forgetting}

Assuming local $L$-smoothness of $\mathcal{R}_{\mathcal{D}_0}$ and an SGD update rule $\omega_{t+1} = \omega_t - \eta g_{\text{task}}(\omega_t)$, a first-order Taylor expansion yields:

\begin{equation}
\begin{aligned}
\Delta_{\text{forget}}
&\approx
-\left\langle
\nabla \mathcal{R}_{\mathcal{D}_0}(\omega_0),
\sum_{t=1}^{T}\eta g_{\text{task}}(\omega_{t-1})
\right\rangle \\
&\quad+
O\!\big(L\sum_{t}\eta^2||g_{\text{task}}(\omega_{t-1})||^2\big).
\end{aligned}
\end{equation}

This expression shows that forgetting arises when the cumulative update has a large projection onto directions that \textbf{increase pretraining loss}. The complete derivation of this result is provided in \cref{app:smoothness}.
If the Style component dominates the update trajectory, the inner product above becomes positive, leading to observable degradation on pretraining-aligned benchmarks even in the absence of semantic conflict.

\subsection{Effect of \methodname}

Let $\mathcal{D}_{\text{self}}$ be the distribution of model-generated samples.
For a mixing ratio $\varepsilon=\lambda/(1+\lambda)$ induced by dataset concatenation,
\begin{equation}
\begin{aligned}
\mathcal{D}_{\text{mix}}
&= (1-\varepsilon)\mathcal{D}_{\text{task}} + \varepsilon\mathcal{D}_{\text{self}}, \\
  g_{\text{mix}}
  &= (1-\varepsilon)g_{\text{task}} + \varepsilon g_{\text{self}}.
\end{aligned}
\end{equation}

Because self-generated text is style-aligned with the pretraining distribution, we have $g_{\text{self}} \approx g_{0}$, and thus
\begin{equation}
\begin{aligned}
\mathrm{Proj}_{\mathcal{S}}(g_{\text{mix}})
  \approx
  (1-\varepsilon)\mathrm{Proj}_{\mathcal{S}}(g_{\text{task}}),
\end{aligned}
\end{equation}
implying that self-mixing \textbf{linearly attenuates} the Style component of the gradient. The complete derivation is provided in \cref{app:mixing}.
Under the same local-smoothness assumption, the corresponding forgetting can be upper-bounded (informally) as:
\begin{equation}
\begin{aligned}
\Delta_{\text{forget}}^{(\text{mix})}\lesssim
(1-\varepsilon)\Delta_{\text{forget}}^{(\text{task})}+O(L\sum_t\eta^2||g_{\text{mix}}||^2),
\end{aligned}
\end{equation}

suggesting that moderate ratios (e.g., $\varepsilon\approx0.5$) optimally suppress style-induced drift while maintaining sufficient task signal. Notably, replacing $\mathcal{D}_{\text{self}}$ with data generated by \textbf{other models} is less effective: their stylistic distribution $S_{\text{other}}$ deviates from $S_0$, resulting in weaker cancellation of style gradients despite potential semantic richness.

\section{Derivations and Additional Details}
\label{app:derivations}

\subsection{Preliminaries and Notation}
Let $\mathcal{R}(\omega) \equiv \mathcal{R}_{\mathcal{D}_0}(\omega)$ denote the pretraining risk,
and let $\{ \omega_t \}_{t=0}^{T}$ be the parameter iterates under SGD on the SFT
distribution $\mathcal{D}_{\text{task}}$ with step size $\eta>0$:
\begin{equation}
\begin{aligned}
&\omega_{t+1} \;=\; \omega_t - \eta\, g_{\text{task}}(\omega_t),\\
&g_{\text{task}}(\omega_t)\;=\;\nabla_{\omega}\,\mathbb{E}_{(x,y)\sim\mathcal{D}_{\text{task}}}
\big[\ell(f_{\omega}(x),y)\big].
\end{aligned}
\end{equation}
We write $\Delta\omega_t := \omega_{t+1}-\omega_t = -\eta\,g_{\text{task}}(\omega_t)$ and
\begin{align*}
    \Delta_{\text{forget}}
\;:=\; \mathcal{R}(\omega_T) - \mathcal{R}(\omega_0).
\end{align*}

\subsection{Local Smoothness and Two-Sided First-Order Remainder}
\label{app:smoothness}
We assume \emph{local $L$-smoothness} of $\mathcal{R}$ around $\omega_0$: for all
$\omega_1,\omega_2$ in a neighborhood of $\omega_0$,
\begin{equation}
\label{eq:Lsmooth}
\|\nabla \mathcal{R}(\omega_1)-\nabla \mathcal{R}(\omega_2)\|
\;\le\; L\,\|\omega_1-\omega_2\|.
\end{equation}
Equivalently, the Taylor remainder is two-sided bounded:
\begin{equation}
\begin{aligned}
\label{eq:taylor-two-sided}
&\big|\,
\mathcal{R}(\omega_2)-\mathcal{R}(\omega_1)
-\langle \nabla \mathcal{R}(\omega_1),\,\omega_2-\omega_1\rangle
\,\big| \\
&\;\le\; \tfrac{L}{2}\,\|\omega_2-\omega_1\|^2.
\end{aligned}
\end{equation}
Applying \eqref{eq:taylor-two-sided} with $\omega_1=\omega_t$ and
$\omega_2=\omega_{t+1}=\omega_t+\Delta\omega_t$ yields the per-step identity with a bounded
remainder:
\begin{equation}
\begin{aligned}
\label{eq:per-step}
&\mathcal{R}(\omega_{t+1})-\mathcal{R}(\omega_t)
\;=\;
\underbrace{\langle \nabla \mathcal{R}(\omega_t),\,\Delta\omega_t\rangle}_{\text{first-order term}}
\;+\;\varepsilon_t, \\
&|\varepsilon_t|\;\le\;\tfrac{L}{2}\,\|\Delta\omega_t\|^2.
\end{aligned}
\end{equation}
Summing \eqref{eq:per-step} over $t=0,\dots,T-1$ gives
\begin{equation}
\begin{aligned}
\label{eq:sum-steps}
&\Delta_{\text{forget}}
\;=\;
\sum_{t=0}^{T-1}\langle \nabla \mathcal{R}(\omega_t),\,\Delta\omega_t\rangle
\;+\;\sum_{t=0}^{T-1}\varepsilon_t,\\
&\Big|\sum_{t=0}^{T-1}\varepsilon_t\Big|
\;\le\; \tfrac{L}{2}\sum_{t=0}^{T-1}\|\Delta\omega_t\|^2.
\end{aligned}
\end{equation}
Using $\Delta\omega_t=-\eta\,g_{\text{task}}(\omega_t)$,
\begin{equation}
\begin{aligned}
\label{eq:sum-main}
&\Delta_{\text{forget}}
\;=\;
-\eta\,\sum_{t=0}^{T-1}\big\langle \nabla \mathcal{R}(\omega_t),\,g_{\text{task}}(\omega_t)\big\rangle
\;+\; \mathcal{E}_1,\\
&|\mathcal{E}_1|\;\le\;\tfrac{L\eta^2}{2}\sum_{t=0}^{T-1}\|g_{\text{task}}(\omega_t)\|^2.
\end{aligned}
\end{equation}
\subsection{Linearization at \texorpdfstring{$\omega_0$}{omega0} and the First-Order Expansion}
\label{app:linearization}
To expose the directional \emph{alignment} driving forgetting, we linearize
$\nabla \mathcal{R}(\omega_t)$ at $\omega_0$:
\begin{equation}
\begin{aligned}
&\nabla \mathcal{R}(\omega_t)
\;=\; \nabla \mathcal{R}(\omega_0) \;+\; \delta_t,\\
&\|\delta_t\|\;\le\; L\,\|\omega_t-\omega_0\|.
\end{aligned}
\end{equation}
Since $\omega_t-\omega_0 = \sum_{j=0}^{t-1}\Delta\omega_j = -\eta \sum_{j=0}^{t-1} g_{\text{task}}(\omega_j)$,
we obtain the bound
\begin{equation}
\label{eq:drift-bound}
\|\delta_t\|
\;\le\;
L\,\eta \sum_{j=0}^{t-1}\|g_{\text{task}}(\omega_j)\|.
\end{equation}
Plugging $\nabla \mathcal{R}(\omega_t)=\nabla \mathcal{R}(\omega_0)+\delta_t$ into
\eqref{eq:sum-main} yields
\begin{equation}
\begin{aligned}
\Delta_{\text{forget}}
\;=\;
&-\eta\,\big\langle \nabla \mathcal{R}(\omega_0),\,\sum_{t=0}^{T-1} g_{\text{task}}(\omega_t)\big\rangle \\
&\;-\;\eta\sum_{t=0}^{T-1}\langle \delta_t,\,g_{\text{task}}(\omega_t)\rangle
\;+\; \mathcal{E}_1.
\end{aligned}
\end{equation}

The drift term is bounded by Cauchy–Schwarz and \eqref{eq:drift-bound}:
\begin{equation}
\begin{aligned}
&\Big|\eta\sum_{t=0}^{T-1}\langle \delta_t,\,g_{\text{task}}(\omega_t)\rangle\Big|
\;\le\; \eta\sum_{t=0}^{T-1}\|\delta_t\|\,\|g_{\text{task}}(\omega_t)\| \\
&\;\le\; L\,\eta^2 \sum_{t=0}^{T-1}\sum_{j=0}^{t-1}\|g_{\text{task}}(\omega_j)\|\,\|g_{\text{task}}(\omega_t)\|.
\end{aligned}
\end{equation}
Under small steps and bounded gradients (the usual local analysis regime), the double sum
is $O\!\big(\sum_t \|g_{\text{task}}(\omega_t)\|^2\big)$. Hence we arrive at the signed
first-order expansion with a \emph{two-sided} remainder:
\begin{equation}
\begin{aligned}
\label{eq:first-order-expansion}
&\Delta_{\text{forget}}
\;=\;
-\big\langle \nabla \mathcal{R}(\omega_0),\,\sum_{t=1}^{T}\eta\, g_{\text{task}}(\omega_{t-1})\big\rangle
\;+\; \mathcal{E}, \\
&|\mathcal{E}|\;\le\; C\,L\,\sum_{t=1}^{T}\eta^2\,\|g_{\text{task}}(\omega_{t-1})\|^2, 
\end{aligned}
\end{equation}

for some absolute constant $C>0$. Equation~\eqref{eq:first-order-expansion} justifies the
main-text use of ``$\approx$'' together with an $O(\cdot)$ remainder: it is an equality up to a
bounded, second-order term whose sign is \emph{a priori} unknown.

\subsection{Effect of Mixing Self-Generated Data}
\label{app:mixing}
Let $\mathcal{D}_{\text{self}}$ be a self-generated distribution (by $f_{\omega_0}$), and define
the mixture
\begin{equation}
\begin{aligned}
&\mathcal{D}_{\text{mix}} \;=\; (1-\varepsilon)\,\mathcal{D}_{\text{task}} \;+\; \varepsilon\,\mathcal{D}_{\text{self}},\\
&g_{\text{mix}}(\omega) \;=\; (1-\varepsilon)\,g_{\text{task}}(\omega)+\varepsilon\,g_{\text{self}}(\omega).   
\end{aligned}
\end{equation}
\paragraph{Assumption (Style--Semantic subspace).}
There exists a subspace $\mathcal{S}$ (Style) and its orthogonal complement
$\mathcal{S}^{\perp}$ (Semantic) such that
$g_{\text{task}}=g_{\text{sty}}+g_{\text{sem}}$ with $g_{\text{sty}}\in\mathcal{S}$,
$g_{\text{sem}}\in\mathcal{S}^{\perp}$. Moreover, self-generated samples are style-aligned with
$\mathcal{D}_0$, so that $\mathrm{Proj}_{\mathcal{S}}\,g_{\text{self}}(\omega_0)\approx 0$.
Then, at initialization,
\begin{equation}
\begin{aligned}
\label{eq:proj-attenuation}
&\mathrm{Proj}_{\mathcal{S}}\,g_{\text{mix}}(\omega_0)
\;=\;(1-\varepsilon)\,\mathrm{Proj}_{\mathcal{S}}\,g_{\text{task}}(\omega_0)\\
&\;+\;\varepsilon\,\mathrm{Proj}_{\mathcal{S}}\,g_{\text{self}}(\omega_0)
\;\approx\; (1-\varepsilon)\,\mathrm{Proj}_{\mathcal{S}}\,g_{\text{task}}(\omega_0). 
\end{aligned}
\end{equation}

Replacing $g_{\text{task}}$ with $g_{\text{mix}}$ in \eqref{eq:first-order-expansion} yields
\begin{equation}
\begin{aligned}
\Delta_{\text{forget}}^{(\text{mix})}
&=
\big\langle \nabla \mathcal{R}(\omega_0),\,
\textstyle\sum_{t=1}^{T}\eta\, g_{\text{mix}}(\omega_{t-1})\big\rangle \\
&\quad + \mathcal{E}^{(\text{mix})}.
\end{aligned}
\end{equation}
Combining \eqref{eq:proj-attenuation} with the subspace assumption shows that the Style
component of the signed main term is reduced by a factor $(1-\varepsilon)$, while the Semantic
component is preserved. Under the same local-smoothness and small-step conditions,
\begin{equation}
\begin{aligned}
\big|\mathcal{E}^{(\text{mix})}\big|
\;\le\; C\,L\,\sum_{t=1}^{T}\eta^2\,\|g_{\text{mix}}(\omega_{t-1})\|^2,
\end{aligned}
\end{equation}
and the Style-driven part of $\Delta_{\text{forget}}^{(\text{mix})}$ scales roughly as $(1-\varepsilon)$
times that of $\Delta_{\text{forget}}^{(\text{task})}$, formalizing the attenuation effect reported in the
main text.

\subsection{Remarks on the Style Perplexity Gap}
\label{app:ppl}
The quantity
\begin{equation}
\begin{aligned}
\Delta_{\mathrm{ppl}}
\;=\; &\mathbb{E}_{x\sim \mathcal{D}_{\text{task}}}\!\big[-\log p_{\omega_0}(x)\big]\\
\;&-\; \mathbb{E}_{x\sim \mathcal{D}_{0}}\!\big[-\log p_{\omega_0}(x)\big]
\end{aligned}
\end{equation}
is a proxy for stylistic out-of-distribution distance measured by $f_{\omega_0}$. Since
$-\log p_{\omega_0}(x)=\sum_{t}-\log p_{\omega_0}(x_t\mid x_{<t})$, it is the standard
sequence NLL under the base model. Empirically, larger $\Delta_{\mathrm{ppl}}$ correlates with
larger $\|\mathrm{Proj}_{\mathcal{S}}\, g_{\text{task}}(\omega_0)\|$, consistent with the view that
style mismatch enlarges the Style component of the update.

\end{document}